# Blind Modulation Classification via Combined Machine Learning and Signal Feature Extraction

Jafar Norolahi, Paeiz Azmi
Department of Electrical and Computer Engineering
Tarbiat Modares University
Tehran, Iran
PAZMI@modares.ac.ir

*Abstract*— In this study, an algorithm to blind and automatic modulation classification has been proposed. It well benefits combined machine leaning and signal feature extraction to recognize diverse range of modulation in low signal power to noise ratio (SNR). The presented algorithm contains four. First, it advantages spectrum analyzing to branching modulated signal based on regular and irregular spectrum character. Seconds, a nonlinear soft margin support vector (NS SVM) problem is applied to received signal, and its symbols are classified to correct and incorrect (support vectors) symbols. The NS SVM employment leads to discounting in physical layer noise effect on modulated signal. After that, a k-center clustering can find center of each class. finally, in correlation function estimation of scatter diagram is correlated with pre-saved ideal scatter diagram of modulations. The correlation outcome is classification result. For more evaluation, success rate, performance, and complexity in compare to many published methods are provided. The simulation prove that the proposed algorithm can classified the modulated signal in less SNR. For example, it can recognize 4-QAM in SNR=-4.2 dB, and 4-FSK in SNR=2.1 dB with %99 success rate. Moreover, due to using of kernel function in dual problem of NS SVM and feature base function, the proposed algorithm has low complexity and simple implementation in practical issues.

## I. INTRODUCTION

In the recent years, automatic modulation classification (AMC) has been widely used to classify the modulation type of the transmitted signal. In fifth generation (5G), the modulation type that is used by the transmitter and the receiver can change frame by frame. In addition, 5G eNode B adaptively selects the appropriate modulation and coding scheme according to the channel state to avoid a given block error rate [1].

In the current 5G and unmanned aerial vehicle (UAV) communications, multipath fading in the system leads to difficulty in AMC because eNode B and users receive and transmit signal from multiple sources and multiple directions [2]. Additionally, in the future of beyond 5G (B5G) communications, due to that massive multiple-input and multiple-output (MIMO) is more interested, it is feeling an AMC is more necessary. So, researching on more efficient algorithms to solving the AMC problem is attended in many literatures [3]–[6].

In general, modulation classifier has two steps, signal preprocessing and classification algorithms. In preprocessing noise reduction, estimation of signal parameters such as carrier frequency and signal power are provided. In the classification algorithms, there are three general categories of classification algorithms, likelihood-based (LB) [7]– [12], feature-based (FB) [13]– [18], and using a machine learning (ML) and artificial neural network (ANN) [19]. The LB compares the likelihood ratio of each possible hypothesis against a threshold, which is derived from the probability density function of the observed wave. Multiple likelihood ratio test (LRT) algorithms have been proposed: Average LRT [20], Generalized LRT [21], and quasi-hybrid LRT [22]. Next, FB method, classification is decided based on several observed features. Also, ML and ANN structures such as multi-layer perceptron (MLP) have been widely used as modulation type classifiers [19] and [23]- [24].

However, one of most advantageous classifiers is support vector machine (SVM) due to its outstanding advantages. One of SVM advantages is utilization of kernel function because cost of practical implementation depends on kernel function. For instance, in [25], a practical SVM classifier was proposed. Its results show that the classifier with kernel function of radical basis function (RBF) costs more than 50000 times power consumption per classification compared to a linear kernel function. Likewise, the local and global SVM classification were conveyed by the aid of LIBSVM library [26].

In this paper, we combined a machine learning scheme with feature extraction to improve current AMC in their advantages. proposed algorithm benefits two techniques in each main parts, the nonlinear soft-margin support vector machine (NS SVM) problem and k-center method.

In NS SVM, we solve a dual problem of SVM to reduce the processing volume and time [27]. The NS SVM well reduce the noise physical channel effect to recognizing of scatter diagram symbols. So, in next step of algorithm, k-center approach can easily estimate the centers of symbols. In final steps of algorithm, estimated scatter diagram is correlated with ideal scatter diagrams of modulation that are per-saved as a n information stock. The simulation results show proof preference in proposed algorithm in compare to other reported algorithms.

This paper is organized as follows, after introduce the suggested algorithm in part I, basis of the proposed method of the proposed method are informed in part II. The part II includes four subsections. In first subsection, the spectrum analysis is illustrated. thenAMC mathematical model based on related mathematic of nonlinear SVM problem is described are presented in second subsection. Next, feature extraction is described in third subsection. Later, the correlation function in fourth subsection.

Moreover, in partIII, proposed algorithm simulation results are provided. furthermore, the simulation results and their analyses are described AMC success rate of modulations and performance. Finally, in last part, the conclusion of this study is explained.

## II. BASIS OF THE PROPOSED ALGORITHM

The proposed method is mainly based on NS SVM problem and k-center method. In this research, modulation recognition totally includes two categories processes, NS SVM problem and features extraction using k-center method. Proposed algorithm is shown in Fig. 1. In fact, in the proposed method, the scatter diagram of received signal is estimated to performing correlation with ideal scatter diagrams of modulation per-saved in an information stock.

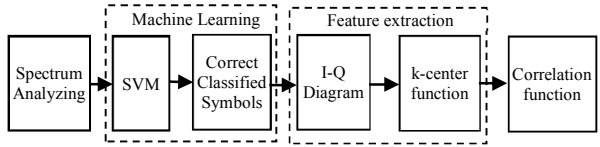

Fig. 1 : The proposed feature-based automatic modulation classification algorithm

As shown in Fig.1, algorithm has totally four parts, spectrum analyzing, machine learning, feature extraction, and correlation function. In first part, the spectrum analyzing is performed to split some modulated signal by their spectrum features. Next, in machine learning part a NS SVM is implemented to received signal. In addition, in feature extraction part, we provide a scatter diagram and k-center function. Finally, fourth part correlates estimated scatter diagram with ideal scatter diagrams of modulations. In following, the all parts are described.

### A. Spectrum analysis

in some studies, such as [28]-[31], spectrum analyzing is conveyed to modulation classification. In this paper, we use an approach for using spectrum analysis in modulation classification that is based on spectrum features. It classifies modulated signal to two groups, regular-shape and irregular-shape spectrum.

Modulated signal with regular-shape spectrums always have a fix and deterministic pattern, while the irregular-shape one does not have any fix or deterministic pattern in its spectrum. For instance, 4FSK modulation always has four peaks in its spectrum while QAM and PSK spectrums are not fix, and their spectrum have several varying peaks with erratic frequency distances. For more clearance, Fig. 2 shows shapes of spectrums for 16QAM, 16PSK, 2FSK, and 4FSK.

So, we can discriminate between modulated signals by estimating and extracting the peaks of the frequency spectrum of the received signal and then comparing the results with the frequency spectrum of each modulation. In continuation of the work, the proposed method classifies irregular-shape spectrum modulated signals group to QAM or PSK subgroups; and considering the number of peaks, it classifies regular-shape spectrum modulation signals to FSK subgroups.

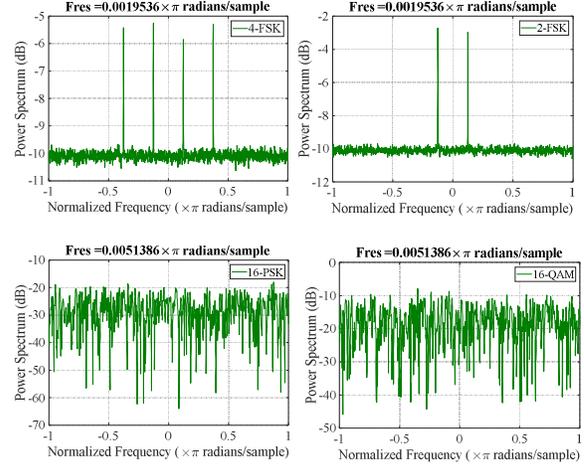

Fig. 2 : The baseband frequency spectrum of 16PSK, 16QAM, and 4FSK in the spectrum analyzer, 16QAM and 16PSK modulated signals have irregular-shape spectrum, and 4FSK modulated signal has regular-shape spectrum.

### B. AMC mathematical model

modulation classification problem is confronted with a noisy data that leads to overlapping classes. In this situation, there are nonlinear separable data and nonlinear decision surfaces. In the case of noisy data and overlapping classes of samples, the soft margin SVM, having slack variables in the problem, is more applicable than finding only maximum margin in hard margin SVM. So, we proposed a solution in this study suggests a soft margin SVM to handle classification of nonlinear separable data.

Assume a set of modulated data examples along $X \in \mathbb{R}^d$ with their corresponding (for example QPSK modulated data) labels $Y$ where $y_n \in \{1,2,3,4\}$. Given a training data set consisting of example–label pairs, $\{(x_1,y_1), \ldots (x_N,y_N)\}$, we would like to estimate parameters of the model that will give the smallest classification error. In this situation, according to [32], optimization problem of soft margin SVM can assumed to:

$$\min_{w, w_0, \{\xi_n\}_{n=1}^N} \frac{1}{2}\|w\|^2 + c\sum_{n=1}^{N} \xi_n$$

$$\text{s.t.} \quad y^{(n)}\left(w^T x^{(n)} + w_0\right) \geq 1-\xi_n \quad n=1,\ldots N \quad (2)$$

$$\xi_n \geq 0$$

where $w$ is a vector normal to the hyperplane, $w_0$ is the intercept, and $\xi_n$ is slack variable corresponding to each

example–label pair $(x_n, y_n)$ that allows a particular example to be within the margin or even on the wrong side of the hyperplane. We subtract the value of $\xi_n$ from the margin, constraining $\xi_n$ to be non-negative. The parameter $C>0$ trades off the size of the margin and the total amount of slack that we have. This parameter is called the *regularization parameter* since the margin term in the objective equation (2) is a regularization term.

Assume a transformation $\phi: \mathbb{R}^d \rightarrow \mathbb{R}^m$ on the feature space. So, we have $x \rightarrow \phi(x)$ and $\phi(x) = [\phi_1(x), \dots \phi_m(x)]$ where the $\{\phi_1(x), \dots \phi_m(x)\}$, is a set of basic functions or features. By substituting of $\phi(x)$ in (1),

$$\min_{w, w_0, \{\xi_n\}_{n=1}^N} \frac{1}{2}\|w\|^2 + C\sum_{n=1}^{N} \xi_n \qquad (3)$$

$$\text{s.t.} \quad y^{(n)}\left(w^T\phi(x^{(n)}) + w_0\right) \geq 1 - \xi_n \quad n=1,\dots,N$$
$$\xi_n \geq 0$$

Where $w \in \mathbb{R}^m$ is the weight set that must be estimated, and in the case of $m \gg d$ (very high dimensional feature space) then there are many more parameters to learn. The equation (3) is known as the primal SVM problem. considering to inputs that $x \in \mathbb{R}^d$ with $d$ features. Since $w$ is of the same dimension as $x$, this means that the number of parameters (the dimension of $w$) of the optimization problem grows linearly with the number of features.

In the following, we consider an equivalent optimization problem (the so-called dual view), which is independent of the number of features. Instead, the number of parameters increases with the number of examples in the training set. This is useful for problems where we have more features than the number of examples in the training dataset. The dual SVM also has the additional advantage that it easily allows kernels to be applied [32]. So, applying proper kernel can decrease cost of practical implementation cost. We call the variables $w$, $w_0$, and $\xi$ corresponding to the primal SVM the primal variables.

We use $\alpha_n \geq 0$ as the Lagrange multiplier corresponding to the constraint (3) that the examples are classified correctly and $\xi_n \geq 0$ as the Lagrange multiplier corresponding to the non-negativity constraint of the slack variable; see (3). The Lagrangian is then given by

$$\mathfrak{L}(w,b,\xi,\alpha,\gamma) = \frac{1}{2}\|w\|^2 + C\sum_{n=1}^{N}\xi_n$$
$$- \sum_{n=1}^{N} \alpha_n \left(y_n(\langle w, x_n \rangle + b) - 1 + \xi_n\right) - \sum_{n=1}^{N} \gamma_n \xi_n \qquad (4)$$

By differentiating the Lagrangian (4) with respect to the three primal variables $w$, $w_0$, and $\xi$ respectively, we obtain

$$\frac{\partial \mathfrak{L}}{\partial w} = w^T - \sum_{n=1}^{N} \alpha_n y_n x_n^T \qquad (5,6,7)$$

$$\frac{\partial \mathfrak{L}}{\partial b} = \sum_{n=1}^{N} \alpha_n y_n$$
$$\frac{\partial \mathfrak{L}}{\partial \xi_n} = C - \alpha_n - \gamma_n$$

substituting results of (5), (6), and (7) in (4) optimization problem of Soft-margin SVM in a transformed space (duality problem) is obtained as follow:

$$\text{Max}_{\alpha} \left\{ \sum_{n=1}^{N} \alpha_n - \frac{1}{2}\sum_{n=1}^{N}\sum_{m=1}^{N} \alpha_n \alpha_m y^{(n)} y^{(m)} \phi(x^{(n)})^T \phi(x^{(m)}) \right\} \qquad (8)$$

Subject to $\sum_{m=1}^{N} \alpha_n y^{(n)} = 0 \quad 0 \leq \alpha_n \leq C \quad n=1, \dots, N$

The (8) equation can be solved by quadratic programming. Additionally, we have inner products $\phi(x^{(n)})^T \phi(x^{(m)})$, only $\alpha = [\alpha_1, \dots, \alpha_n]$ needs to be learnt. In other words, it is not necessary to learn $m$ parameters as opposed to the primal problem. Because of that $\phi(x^{(m)})$ is maybe non-linear function, it is possible to provide a non- linear classifier in examples $x_n$.

Since $\phi(x^{(m)})$ could be a non-linear function, we can use the SVM (which assumes a linear classifier) to construct classifiers that are nonlinear in the examples $x_n$. So, we can separate data set nonlinearly. It gives us an advantage to simplify inner product in the dual SVM, $\phi(x^{(n)})^T \phi(x^{(m)})$. thus, Instead of clearly defining a non-linear feature map and computing the resulting inner product between examples $\phi(x^{(n)})^T$ and $\phi(x^{(n)})^T$, we define a similarity function $K(x^{(n)}, x^{(m)})$ between $x^{(n)}$ and $x^{(m)}$ kernel. The function $K$ is named kernel function.

As a result, in this step (having an inner products), kernel SVM can help to decrease the processing of inner products (deceasing the practical implementation cost) of $\phi(x^{(n)})^T \phi(x^{(m)})$.

We use Gaussian or Radial Basis Function (RBF) to computing of inner products of $\phi(x^{(n)})^T \phi(x^{(m)})$ without any working on mapped data, $\phi(x)$. so, we can show the kernel function as

$$\phi(x^{(n)})^T \phi(x^{(m)}) = k(x^{(n)}, x^{(m)}) \qquad (9)$$

where $K$ is RBF kernel is calculated by (10).

$$k(x^{(n)}, x^{(m)}) = exp\left(-\frac{\|x^{(n)} - x^{(m)}\|^2}{\gamma}\right) \qquad (10)$$

Substituting (10) in (8) yields a new optimization problem that It can indeed be efficiently computed, with a cost proportional to $d$ (the dimensionality of the input) instead of $m$.

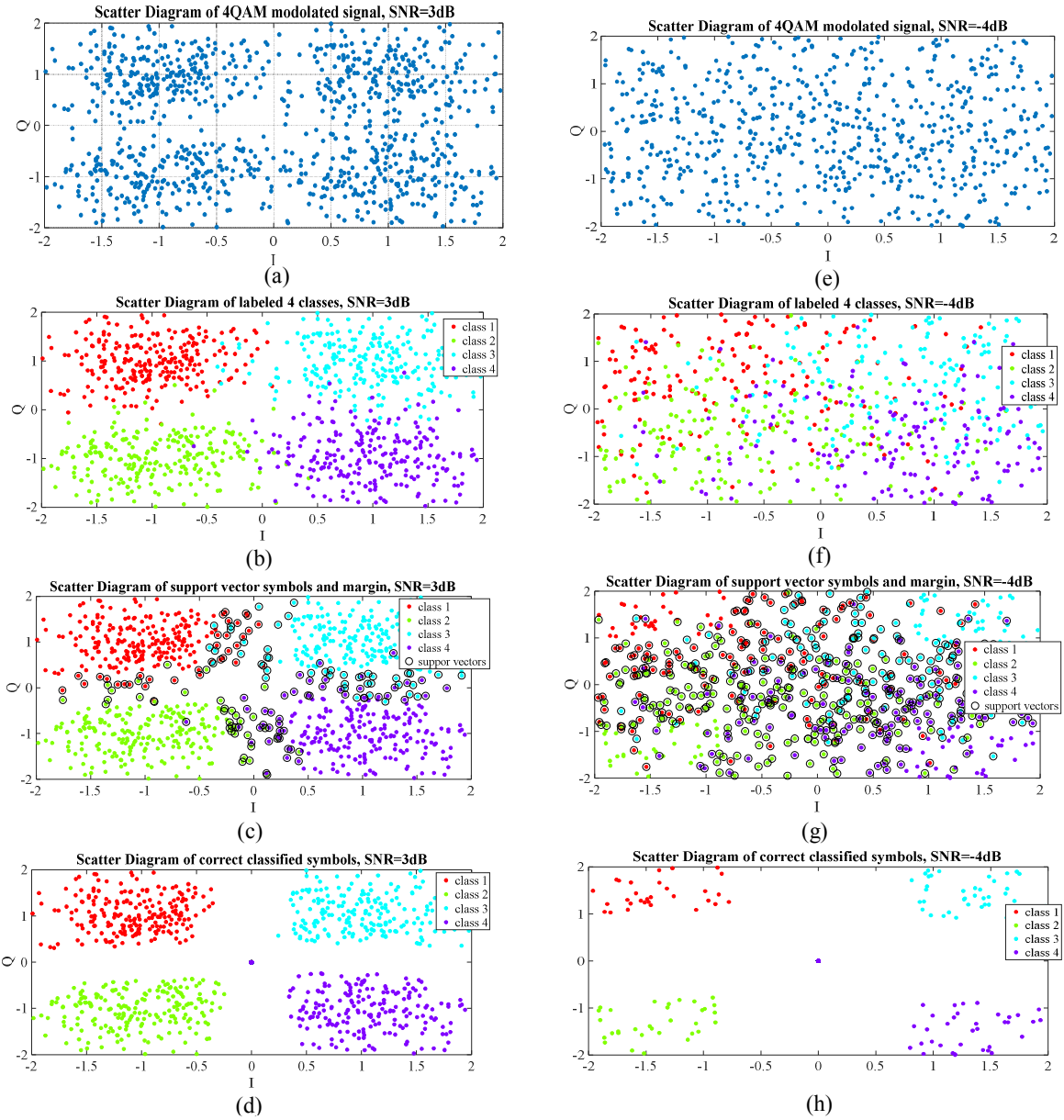

Fig. 3: An example of suggested classifier to QAM recognition

$$\text{Max} \left\{ \sum_{n=1}^{N} \alpha_n - \frac{1}{2} \sum_{n=1}^{N} \sum_{m=1}^{N} \alpha_n \alpha_m \boldsymbol{y^{(n)}} \boldsymbol{y^{(m)}} K(\boldsymbol{x^{(n)}}, \boldsymbol{x^{(m)}}) \right\}_\alpha \quad (11)$$

Subject to $\quad \sum_{m=1}^{N} \alpha_n \boldsymbol{y^{(n)}} = \boldsymbol{0}$

Now, optimization of (11) can be solved to obtain classifying data.

$$\hat{y} = sign(\boldsymbol{w_0} + \boldsymbol{w}^T \phi(\boldsymbol{x})) \quad (12)$$

where $w = \sum_{\alpha_n > 0} \alpha_n y^{(n)} \phi(x)$

and $\boldsymbol{w_0} = \boldsymbol{y^{(s)}} - \boldsymbol{w^T} \phi(\boldsymbol{x^{(s)}})$

$$\hat{y} = sign(\boldsymbol{w_0} + \sum_{\alpha_n > 0} \alpha_n y^{(n)} k(\boldsymbol{x^{(n)}}, \boldsymbol{x^{(m)}}))$$

$$\boldsymbol{w_0} = \boldsymbol{y^{(s)}} - \boldsymbol{w}^T \sum_{\alpha_n > 0} \alpha_n y^{(n)} k(\boldsymbol{x^{(n)}}, \boldsymbol{x^{(m)}}))$$

Subsequently, Fig. 3 shows how a modulated is recognized by the applied NS SVM classifier in SNR=3dB and SNR= -

4dB. As it shows, Fig. 3.a indicates the scatter diagram of 4QAM modulated signal symbols in SNR=3dB. Because there are 4 classes in modulated signal, in Fig. 3.b the labeled symbols of modulated signal are appeared by 4 different colors. Next, in Fig. 3.c, the results of NS SVM to classify 4-classes symbols, including correct classified and incorrect classified symbols (support vectors), are exposed. The correct classified symbols of each class is indicated by color dots without black circle, and incorrect classified symbols are classified by black circles around color dots. In Fig. 3.d, the scatter diagram of correct classified symbols is presented. Similarly, for more illustration, Fig. 3. e, f, g, h well demonstrate the NS SVM processing to 4QAM modulated signal in SNR= -4dB. Whatever is notable in Fig. 3 is that if the Fig. 3.c and Fig. 3.g are compared, it seems the number of support vectors (incorrect classified symbols) is increased in less SNR=-4dB in compare to more one SNR=3dB. It also leads to decreasing of correct classified symbols of each class in Fig. 3.h in compare to Fig. 3.d. However, in both of them the 4QAM modulation is symbols are classified successfully. Subsequently, Fig. 3.h in compare to Fig. 3 well shows that the proposed NS SVM can excellently discount the physical channel noise effect on symbols.

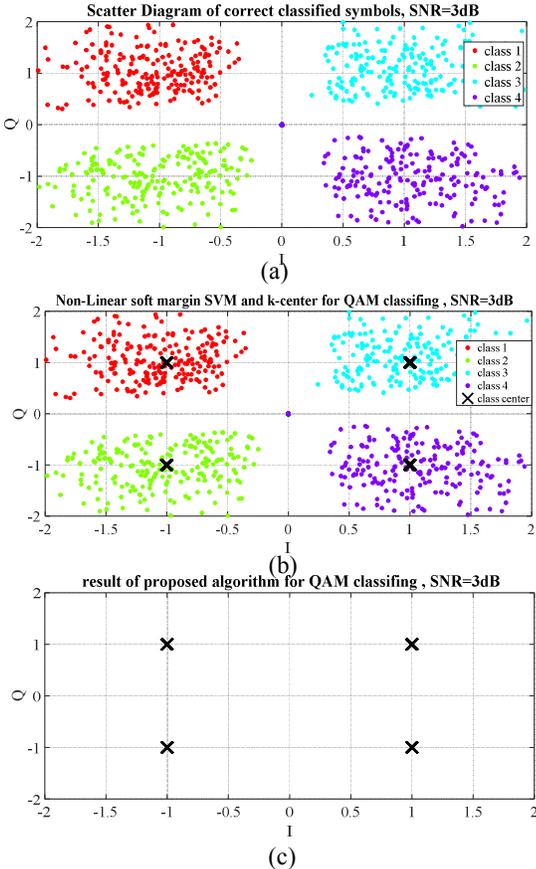

Fig. 4: Implementation of k-center in algorithm to NS SVM correct classified symbols at SNR=3dB.

## C. Feature extraction

In this step, we provide a correlation between estimated scatter diagram of received signal (Fig. 3.h and Fig. 3.d) and ideal scatter diagram of pre-saved modulations. We use k-center algorithms to extract the scatter diagram of the signal in low SNRs [33]. Similar to Fig. 3, an example for using this method for 4QAM modulation with SNR= 3 dB is shown in Fig. 4 where Fig. 4 .a indicates the correct classified symbols by NS SVM classifier (Fig. 3.d) and in Fig. 4 .b the multiplication signs, (×), show each class center that estimated by using k-center algorithms. Considering the estimated centers in Fig. 4.c as a scatter diagram of classified signal, the result of k-center method is proper to correlation performing with pre-saved ideal scatter diagrams of diverse range of modulations.

## D. Correlation function

In this step, the k-center algorithm result as estimated scatter diagram of received signal is correlated with ideal scatter diagram of modulations that have been saved perversely. The correlation result shows amount of matching between estimated scatter diagram and pre-saved ideal scatter diagram of modulations. If we name $\mathcal{C}$ the correlation, we have $0 \leq \mathcal{C} \leq 1$. It is absolutely clear; the most achieved correlation recognizes the classified modulation. In TABLE I, the results of QAM classifying is informed in SNR=3dB.

TABLE I :  results of QAM classifying in SNR=3dB

| index | Estimated scatter diagram received signal | Pre-saved scatter diagram of ideal modulation | Correlation results ($\mathcal{C}$) |
|---|---|---|---|
| 1 | QAM | QAM | 0.935 |
| 2 | QAM | 16QAM | 0.235 |
| 3 | QAM | 64QAM | 0.096 |
| 4 | QAM | BPSK | 0.112 |
| 5 | QAM | QPSK | 0.236 |
| 6 | QAM | 8PSK | 0.152 |
| 7 | QAM | 2FSK | 0.112 |
| 8 | QAM | 4FSK | 0.236 |

Moreover, related to SNR, a threshold can be adjusted to correlation levels. So, in TABLE II, our suggested threshold for some modulation are reported. For example, the adjusted threshold for SNR=3dB is 0.8.

TABLE II : suggestable threshold adjustment to modulation classification in correlation function for SNR=3dB.

| Pre-saved scatter diagram of ideal modulation | suggestable threshold to Correlation ($\mathcal{C}$) |
|---|---|
| 4QAM | 0.8 |
| 16QAM | 0.8 |
| 64QAM | 0.9 |
| BPSK | 0.8 |
| QPSK | 0..8 |
| 8PSK | 0.85 |
| 2FSK | 0.8 |
| 4FSK | 0.8 |

## III. PROPOSED ALGORITHM SIMULATION RESULTS

In this section, comparison results are presented in two parts. First part is dedicated to present success rate of the proposed algorithm in classification of the modulations, and second part is focused in comparing the performance of the proposed method with recent published modulation classifiers.

One of the important criteria of modulation classifiers is success rate. Success rate presents the reliability of the classification. Different modulated signals have different features in their signaling, constellation, and geometrical construction. In this situation, difficulty of features extraction is not equal in all type of modulation, and it is related to SNR. In other words, there is a trade-off between difficulty and SNR. Moreover, required SNR to achieve specific success rate is different in each modulation. So, different modulations need different SNRs to obtain each success rate.

In this section, we aimed two success rate levels of 99% and 80%, in order to evaluation of the proposed algorithm.

Fig. 5 shows the success rates of classification in terms of SNR. These curves are analyzed in TABLE III. This analysis includes SNR ranges for 99% and 80% success rates. In addition, number of symbols and number of iterations are set $10^3$ and $10^5$, respectively. Where number of symbols indicates the number of symbols that we received from the modulated signal, and iterations are number of the proposed algorithm repetition to classify modulated signal. Additionally, equivalent symbol number provides similar conditions in the modulated signals sampling. In the other hand, the large number of repetition (iteration), $10^5$ for any classification, can improve the reliability of classifier results. Iteration does not have any influences on the success rate, but it can improve the reliability and precise of achieved success rate.

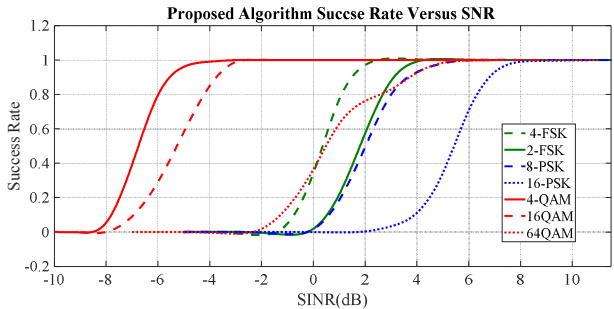

Fig. 5: Success rate of proposed algorithm in terms of SNR for different modulations.

As TABLE III shows, the proposed method achieves 99% success rate in SNR as low as -4.2dB for 4QAM, and -3.1dB for 16QAM modulations. Results also show that 99% success rate can happen in SNR=5.5dB for 64QAM.

Moreover, the worst case for 99% success rate is for 16-PSK modulation that is detectable by success rate of 99% in SNR=7.8dB. Moreover, in 99% success rate, 8PSK is detected at SNR=5.5dB. furthermore, Success rate results show that achieving a success rate of 99%for classifying of 2FSK and 4FSK need to SNR=4dB and 2.1dB respectively. Subsequently, higher orders of modulations would need higher SNRs to achieve the same success rates, except in FSK modulation that 4FSK can be detected with about 2dB lower SNR than 2FSK.

In addition, 80% success rate levels in order to more evaluation of the proposed algorithm can be considered in TABLE III. Finally, it can be said that required SNRs are different for the sake of different difficulties in features extraction.

TABLE III : Success rates of the proposed algorithm at different SNRs

| Modulation type | 99% success rate | 80 % success rate | symbols | iteration |
|---|---|---|---|---|
| 64-QAM | 5.5 dB | 2.2 dB | $10^3$ | $10^5$ |
| 16-QAM | -3.1 dB | -4.2 dB | $10^3$ | $10^5$ |
| 4-QAM | -4.2 dB | -6 dB | $10^3$ | $10^5$ |
| 4-FSK | 2.1 dB | 1 dB | $10^3$ | $10^5$ |
| 2-FSK | 4 dB | 2.8 dB | $10^3$ | $10^5$ |
| 8-PSK | 5.5 dB | 3 dB | $10^3$ | $10^5$ |
| 16-PSK | 7.8 dB | 6.5 dB | $10^3$ | $10^5$ |

In the continuation of this section, we compared the proposed classifier with other classifiers in different SNR for 16QAM to success rates and complexities. Results in TABLE IV is provided according to whatever is reported in [34]-[38]. Because of difference between studied methods in TABLE IV , their results can be different obviously. For example, AMC with ELM [35] method uses an extreme learning machine as a classifier, which has a faster learning process and better performance than conventional machine learning method. But, it needs SNR=7dB to achieve 99% success rate while suggested algorithm needs to SNR=-3.1dB to that.

In addition, in [36] a method based on genetic algorithm has been proposed that it needs SNR=15dB for success rate of 99%. In [37] a feature-based classifier has been suggested that has success rate about 99% with SNR=11dB. Additionally, a method for classifying the electromagnetic signals of a radar or communication system according to their modulation characteristics has been presented in [38]. It identifies 16QAM with success rate of 99.26% and SNR=30dB.

Furthermore, Other comparison that includes complexities of AMCs is also presented in TABLE IV according to whatever is reported in [34]-[38]. Complexity clearly means the volume of mathematics and number of steps required to recognize modulation types.

As mentioned earlier, LB methods, despite generating optimal responses, suffer from greater computational complexity than feature-based modulation methods due to the need for prior knowledge of the received signal [39]-[41]. TABLE IV indicates two groups of complexities based on reported results of the mentioned references. First group is low complexity algorithms includes proposed algorithm, AMC with ELM [35], GPOS [36], FB [37], AMC with HMLN [38]. Second group is acceptable complexity that includes ALRT, Quasi-ALRT, Cumulant-based, Quasi-HLRT [34].

In summary, results of TABLE IV confirm that proposed AMC is belonged to low complexity and low SNR group AMCs.

TABLE IV : Performance comparison with other automatic classifiers.

| Classification algorithm | SNR (dB) | Success rate | complexity | Modulation type |
|---|---|---|---|---|
| Proposed algorithm | -3.1 | 99% | Low | 16-QAM |
| ALRT [34], L=1, ηA=1 | 7 | 99% | Acceptable [34] | 16-QAM |
| Quasi-ALRT [34] | 30 | 88% | Acceptable [34] | 16-QAM |
| HLRT [34], µ H not specified | 9 | 99% | Acceptable [34] | 16-QAM |
| Cumulant-based [34], Nm=2, µH=−0.68 | 9 | 99% | Acceptable [34] | 16-QAM |
| Quasi-HLRT [34], threshold = 1 | 19 | 99% | Acceptable [34] | 16-QAM |
| AMC with ELM [35] | 7 | 99 % | Low [35] | 16-QAM |
| GPOS [36], symbol length 4096 | 15 | 99% | Low [36] | 16-QAM |
| FB [37] | 11 | 99% | Low [37] | 16-QAM |
| AMC with HMLN [38] | 30 | 99.26% | Low [38] | 16-QAM |

IV. CONCLUSION

In this study, a new algorithm to blind and automatic modulation classification has been proposed. It well benefits combined NS SVM and k-center method. The presented algorithm contains some processes. First, it advantages spectrum analyzing to branching modulated signal based on regular and irregular spectrum character. Seconds, NS SVM problem is applied to received signal, and its symbols are classified to correct and incorrect (support vectors) symbols. The NS SVM employment leads to discounting in physical layer effect on modulated signal. After crossing the support vectors out from classified symbols, a k-center method can center each class of classified symbols. The outcome of k-center is an estimation of scatter diagram of received signal that is correlated with diverse range of pre-saved ideal scatter diagram of modulations. in continuation of our study, the simulation results are provided to indicate success rate, performance, and complexity in compare to many published methods. The simulation prove that the proposed algorithm can classified the modulated signal in less SNR and lower complexity.